\documentclass[10pt,twocolumn,letterpaper]{article}

\usepackage[pagenumbers]{cvpr} %

\definecolor{cvprblue}{rgb}{0.21,0.49,0.74}
\usepackage[pagebackref,breaklinks,colorlinks,allcolors=cvprblue]{hyperref}
\usepackage{booktabs} %
\usepackage{multirow} %
\usepackage{amssymb}  %
\usepackage{array}
\usepackage{graphicx}
\usepackage{subcaption}
\usepackage{float}
\usepackage{stfloats}

\title{Learning Plug-and-play Memory for Guiding Video Diffusion Models}

\author{
Selena Song$^{1*}$ \quad
Ziming Xu$^{1*}$ \quad
Zijun Zhang$^{1}$ \quad
Kun Zhou$^{1\dagger}$ \\
Jiaxian Guo$^{2}$ \quad
Lianhui Qin$^{1}$ \quad
Biwei Huang$^{1}$\\[0.5em]
$^{1}$University of California, San Diego \quad
$^{2}$The University of Tokyo\\[0.5em]
{\tt franciskunzhou@gmail.com}\\
}

\begin{document}
\maketitle
\begingroup
\renewcommand\thefootnote{}\footnotetext{* Equal contribution, listed in alphabetical order.}
\renewcommand\thefootnote{}\footnotetext{$\dagger$ Corresponding author.}
\endgroup
\begin{abstract}
Diffusion Transformer~(DiT) based video generation models have recently achieved impressive visual quality and temporal coherence, but they still frequently violate basic physical laws and commonsense dynamics, revealing a lack of explicit world knowledge. In this work, we explore how to equip them with a plug-and-play memory that injects useful world knowledge. 
Motivated by in-context memory in Transformer-based LLMs, we conduct empirical studies to show that DiT can be steered via interventions on its hidden states, and simple low-pass and high-pass filters in the embedding space naturally disentangle low-level appearance and high-level physical/semantic cues, enabling targeted guidance. Building on these observations, we propose a learnable memory encoder DiT-Mem, composed of stacked 3D CNNs, low-/high-pass filters, and self-attention layers. The encoder maps reference videos into a compact set of memory tokens, which are concatenated as the memory within the DiT self-attention layers.
During training, we keep the diffusion backbone frozen, and only optimize the memory encoder. It yields a rather efficient training process on few training parameters (\eg 150M) and 10K data samples, and enables plug-and-play usage at inference time.
Extensive experiments on state-of-the-art models demonstrate the effectiveness of our method in improving physical rule following and video fidelity.
Our code and data are publicly released here \url{https://thrcle421.github.io/DiT-Mem-Web/}.
\end{abstract}
    
\section{Introduction}
\label{sec:intro}
Recent advances in diffusion-based generative models have dramatically improved the visual quality and temporal coherence of video generation\citep{ho2022video,ho2022imagen}. Built on the diffusion Transformer (DiT) architecture~\cite{DiT}, recent models~\cite{wan2025wan,yang2024cogvideox,kong2024hunyuanvideo} leverage scaling in both model size and training data, achieving remarkable capability in producing high-fidelity video clips.
Yet, generating truly realistic videos remains a major challenge. Beyond sharp frames and smooth motion, videos must obey the same physical laws and commonsense dynamics that govern the real world. 
In practice, even state-of-the-art models often violate basic physics and commonsense, \eg objects passing through solid walls, liquids flowing upward, or collisions that leave both objects completely unaffected.
Such failures highlight the broader issue that current video diffusion models still lack sufficient world knowledge to reliably enforce physical and commonsense consistency during generation.

Equipping video diffusion models with rich world knowledge remains a fundamental yet challenging problem. Existing efforts are predominantly training-based and follow two main strategies. One line of work scales up model capacity and training data in the hope that broad exposure to diverse scenes allows the model to implicitly internalize world knowledge regularities \cite{peebles2023scalable}. Another line introduces additional constraints or supervision into the training objective \cite{yuan2023physdiff,raissi2019physics,liu2024physgen}. However, both strategies substantially increase training cost and complexity, and still struggle to comprehensively cover the vast and diverse knowledge required to handle real-world scenarios \citep{kang2024far}.

In this paper, we aim to develop an efficient solution to equip video diffusion models with a memory that provides useful world knowledge to guide generation. Inspired by the success of in-context memory for Transformer-based LLMs~\cite{brown2020language,lewis2020retrieval,borgeaud2022improving,wu2022memorizing}, we consider a plug-and-play memory design for DiT-based video generation models. 
To this end, we conduct preliminary experiments to study \emph{whether we can guide DiT through in-context intervention}.
We retrieve a few relevant videos, encode them into embeddings, and then inject these embeddings into the hidden states of the DiT during inference. Surprisingly, we find that performing low-pass and high-pass filtering~\cite{cai2021frequency,koo2024flexiedit} in the embedding space can guide the DiT to generate the desired objects and to better follow physical rules, respectively. It suggests that DiT supports guidance via interventions on its hidden states, and straightforward filtering in the frequency domain can disentangle low-level and high-level concept embeddings that effectively steer the generation process.

Building on these findings, we propose \textbf{DiT-Mem}, a learnable, general-purpose memory module that can be plugged into existing video diffusion models to provide guidance during generation. Specifically, we design a lightweight memory encoder composed of stacked 3D CNN layers for spatiotemporal downsampling, high-pass and low-pass filters to extract disentangled high-level (semantic/dynamic) and low-level (appearance/texture) features, and a self-attention layer for feature aggregation. Given a reference video, DiT-Mem compresses it into a small set of memory embeddings, which can be concatenated with hidden states from self-attention layers of the DiT as in-context memory tokens. During training, we only update the parameters of the memory encoder while keeping the DiT backbone frozen, resulting in a parameter-efficient learning process and a truly plug-and-play memory that can be used to guide generation at inference time.

In practice, we only need to finetune few training parameters (\eg 150M) on 10K data samples, leading to a rather efficient training recipe to obtain the general memory for fixed DiT-based video diffusion models. To demonstrate the effectiveness of our approach, we integrate the proposed memory module into state-of-the-art open-source models (\eg Wan2.1 T2V-1.3B~\cite{wan2024video} and Wan2.2 TI2V-5B~\cite{wan2025wan}) and evaluate them on a diverse suite of video generation benchmarks. Across multiple evaluation settings covering object controllability, physical consistency, and long-horizon dynamics, our method yields improvements in visual quality and rule compliance, even surpassing closed-source models (\eg Pika~\cite{pika2024} and Kling~\cite{kling2024}).

\section{Preliminary}
\label{sec:preliminary}

\paragraph{DiT-based Video Generation Model}
We build on a text-to-video (T2V) generation model where a DiT serves as the backbone. 
The DiT-based video generation model typically consists of a text encoder, a vision Transformer~(ViT), and a pre-trained variational autoencoder~(VAE) \cite{rombach2022high}.
Given a natural-language prompt $p$ as input, the text encoder first produces a conditioning embedding $e(p)$. 
At inference time, starting from randomly sampled pure noise in the latent space, the ViT iteratively denoises the latent noise under the guidance of the text condition. The final clean latent $z_0$ is decoded by the VAE into the output video $\hat{v} = \mathcal{D}(z_0)$ that realizes the semantics and dynamics described by the input prompt.
During training, a ground-truth video $v$ is encoded by the VAE into a latent tensor $z = \mathcal{E}(v)$, which is further partitioned into a sequence of spatio-temporal latent patches $x_0$ that act as input token embeddings for the ViT. A forward diffusion process progressively corrupts these tokens with Gaussian noise to obtain $x_t$ at the $t$-th timestep. The noisy latent tokens, the timestep embedding, and the conditioning tokens are concatenated, and fed into the ViT to predict the injected noise $\epsilon_\theta(x_t, t, c)$. It is optimized with a standard denoising diffusion objective \cite{ho2022video}, minimizing the expected L2-distance between the true noise $\epsilon$ and the prediction $\epsilon_\theta(x_t, t, c)$:
\begin{equation}
\mathcal{L}=\mathbb{E}_{t,\,x_0,\,\epsilon,\,c}\left[\left\|\epsilon-\epsilon_\theta(x_t,t,c)\right\|_2^2\right].   
\end{equation}

\paragraph{Problem Statement.}
In this paper, we study the task of training a memory encoder to enhance a pre-trained DiT-based T2V generation model with external world knowledge. We assume access to an external video memory bank composed of captioned reference clips, where each entry is a pair $\langle v_i, c_i \rangle$ of a video $v_i$ and a short description $c_i$. Following a retrieval-augmented generation (RAG) formulation, this memory bank plays the role of an external knowledge corpus. Concretely, a retrieval module first selects a small set of videos that are semantically aligned with the input prompt, and the generator then conditions on both the prompt and the retrieved videos. In practice, for each input natural language prompt $p$, we utilize a semantic encoder to compute its similarity with all the video captions from the memory, recall a set of the most relevant entries, and use a memory encoder to map each retrieved video into a set of continuous embeddings. The resulting memory tokens are concatenated as auxiliary context and injected into the DiT-based model, effectively sampling the target video $v$ from $p(v \mid p, {m_t})$ so that the generated content better realizes the scene and dynamics described by $p$. Throughout both training and inference, the DiT backbone remains frozen, and only the memory encoder is optimized, so that no adaptation of the diffusion model to the memory is required.

\begin{figure*}[t]
  \centering

  \begin{subfigure}{\textwidth}
    \centering
    \includegraphics[width=.98\textwidth]{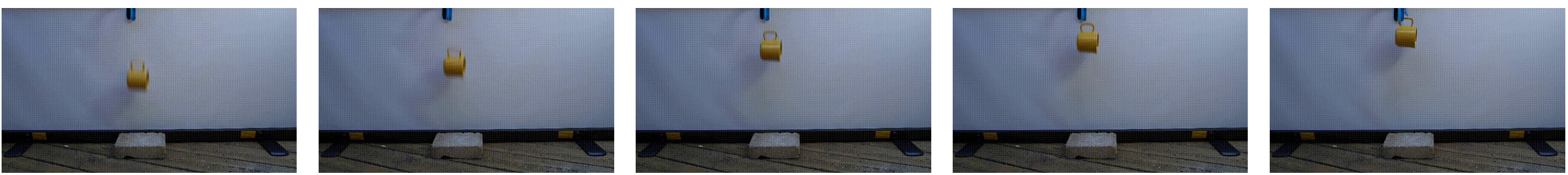}
    \caption{\textbf{Baseline (no steering)}: The mug \textcolor{red}{floats upward, violating basic physical rule about gravity.}}
    \label{fig:case1}
  \end{subfigure}

  \vspace{2pt}

  \begin{subfigure}{\textwidth}
    \centering
    \includegraphics[width=.98\textwidth]{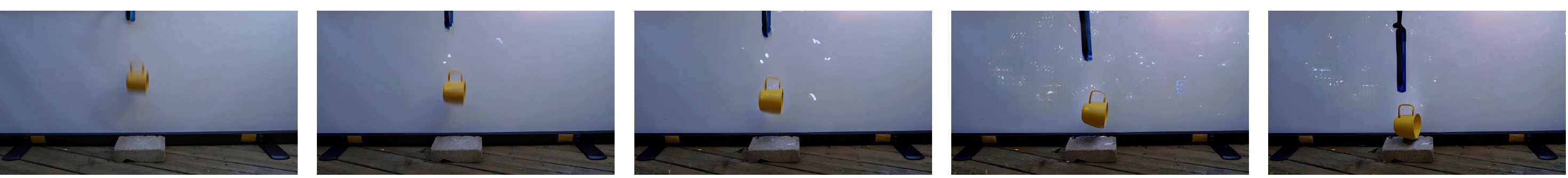}
    \caption{\textbf{Steering with High-Pass Vector}: The mug \textcolor{blue}{naturally falls down}, with a trajectory \textcolor{blue}{following physical rule} but \textcolor{red}{containing background noise}.}
    \label{fig:case2}
  \end{subfigure}

  \vspace{2pt}

  \begin{subfigure}{\textwidth}
    \centering
    \includegraphics[width=.98\textwidth]{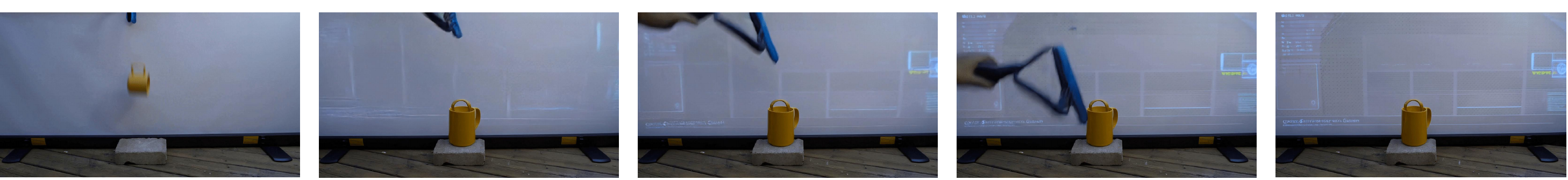}
    \caption{\textbf{Steering with Low-Pass Vector}: The mug \textcolor{red}{suddenly falls down with an unreal turning}, with \textcolor{blue}{new objects introduced} from the source video.}
    \label{fig:case3}
  \end{subfigure}

\caption{\textbf{Qualitative results of frequency-based steering interventions on Wan2.1 I2V-14B.}
  All three cases are generated from the same prompt:
\emph{``A yellow mug is held by a grabber tool in front of a white projection screen with a concrete brick positioned beneath it. The grabber releases the mug. Static shot with no camera movement.''}
Case (b) shows that high-pass features guide physical dynamics.
Case (c) shows that low-pass features encode structural and object-level information.  \textcolor{blue}{Blue} and \textcolor{red}{red} fonts denote positive and negative findings.
}
\vspace{-1mm}
\label{fig:steer_vector}
\end{figure*}

\section{Empirical Study}
\label{sec:empirical}
To explore the effectiveness of equipping a memory module to guide DiT inference (analogous to in-context intervention in LLMs~\cite{brown2020language,lewis2020retrieval}), we conduct an empirical study to examine: (1) whether DiT can respond to guidance through in-context interventions, and (2) whether such guidance can be steered toward specific targets. Inspired by steering-vector-based LLM interventions~\cite{zou2023representation,rimsky2024steering,stolfo2024improving} and frequency-domain feature analysis methods~\cite{cai2021frequency,koo2024flexiedit}, we propose a training-free intervention approach to guide DiT-based video generation models. We evaluate its effectiveness on samples involving physical rule following, assessing how well the intervention steers DiT’s generative behavior toward the desired dynamics.

\subsection{Inference-time Intervention Method}
To examine whether DiT supports guidance from in-context memory, we propose an inference-time intervention method composed of two key components: steering vector extraction and frequency-domain feature filtering. This approach enables training-free control of DiT-based video generation models by injecting semantically meaningful directional signals into the model’s internal representations.

\paragraph{Steering Vector Extraction.}
We first extract steering vectors that encode the conceptual direction of a desired physical property. For each target concept, we generate paired samples consisting of: (1) \emph{positive} variants prompted with physics-consistent descriptions, and (2) \emph{negative} variants prompted with empty or neutral text. During generation, we record the cross-attention outputs from all DiT layers across all denoising timesteps.
For each sample, the spatial patch outputs are averaged to form a single feature vector per layer and timestep. We then compute mean representations for all positive and negative samples, respectively, and define the steering vector as their difference:

\begin{equation}
\mathbf{s}_{t,\ell} \;=\; \mathbb{E}_{j,k}\!\left[\mathbf{V}^{\mathrm{pos}(j)}_{t,\ell,k}\right] \;-\; \mathbb{E}_{j,k}\!\left[\mathbf{V}^{\mathrm{neg}(j)}_{t,\ell,k}\right].
\end{equation}
We denote by $\mathbf{V}_{t,\ell,k}^{\mathrm{pos}(j)}, \mathbf{V}_{t,\ell,k}^{\mathrm{neg}(j)} \in \mathbb{R}^d$ the cross-attention value vectors at denoising timestep $t$, DiT layer $\ell$, spatial position $k$, and sample $j$, generated with the physics-describing prompt and the empty prompt, respectively.
For brevity, we may write $\mathbf{V}_{t,\ell,k}^{(j)}$ when the condition is clear.
The expectation $\mathbb{E}_{j,k}[\cdot]$ is taken over samples $j$ and spatial positions $k$.

\vspace{-2mm}
\paragraph{Frequency-Domain Filtering.}
Next, we process the temporal sequence of steering vectors to isolate meaningful components in the frequency domain.
For each layer $\ell$, the per-timestep steering sequence is concatenated along the time axis.
We then perform low-pass and high-pass filtering on this sequence to disentangle low-level and high-level features.
Concretely, we transform the features into the frequency domain via a 1D Fast Fourier Transform (FFT), and multiply them with the low-pass mask $\mathbf{M}_{\mathrm{low}}$ and high-pass mask $\mathbf{M}_{\mathrm{high}}$ to attenuate high-frequency and low-frequency components, respectively.
According to existing work \cite{cai2021frequency, koo2024flexiedit}, low-frequency and high-frequency features typically capture object-level semantics and rapid motions, respectively, which allows for targeted guidance of object appearance and physical dynamics.
After that, we perform the inverse FFT to transform the filtered features back into the time domain, yielding the filtered steering vectors, which we denote as $\mathbf{s}'_{t,\ell}$.
We further normalize the selected filtered steering vector at each $(t,\ell)$ to obtain $\tilde{\mathbf{s}}_{t,\ell}$.

During inference, the selected steering vector is injected into the corresponding layer $\ell$ at each timestep $t$ via a scaled addition:
\begin{equation} 
\mathbf{V}_{t,\ell,k} \leftarrow \mathbf{V}_{t,\ell,k} + \alpha\, \tilde{\mathbf{s}}_{t,\ell},
\end{equation}
where $\alpha$ is a scaling coefficient to control the guidance strength.
To maintain fine-grained visual quality, the intervention is applied only during the early and mid denoising stages, skipping the final one-third of timesteps.

\subsection{Experimental Results and Findings}

\paragraph{Experimental Setup.}
We conduct experiments on the Wan2.1 I2V-14B model~\cite{wan2024video} to evaluate whether DiT supports guidance from steering vectors derived through in-context interventions.
We collect a small set of samples that explicitly follow specific physical rules (\eg gravity, collision). For each rule, we generate pairs of \emph{positive} videos using physics-consistent prompts and \emph{negative} videos using empty or neutral prompts, and manually filter out false positives that fail to follow the intended rule.
Using these curated pairs, we compute steering vectors and apply low-pass and high-pass frequency filters respectively, then use the resulting filtered signals to guide video generation.
\vspace{-3mm}
\paragraph{Results and Findings.}
As shown in Fig.~\ref{fig:steer_vector}, our experiments yield three key observations:
\begin{itemize}
    \item \emph{DiT responds effectively to steering guidance.} Injecting steering vectors during inference successfully influences the model’s generative behavior, confirming that DiT can be modulated through in-context interventions.
    \item \emph{Low-pass filtered features capture object-level information.} Steering with low-frequency components preserves global scene structure and object identity, resulting in more coherent appearance control.
    \item \emph{High-pass filtered features reinforce physical consistency.} High-frequency components predominantly affect motion and dynamics, leading to trajectories that more accurately follow physical laws such as gravity and inertia.
\end{itemize}

\vspace{-3mm}
\paragraph{Summary.}
These results demonstrate that DiT can indeed respond to guidance from in-context interventions and that such guidance can be steered toward targeted objectives by manipulating frequency-filtered feature components. This finding highlights the promise of using frequency-domain filtering on encoded video features to obtain targeted low- and high-level knowledge as a form of memory guidance. However, as the current training-free intervention still introduces noise and occasional distortions, we will focus on developing efficient training strategies to stabilize and enhance controllable video generation.

\begin{figure*}[t]
  \centering
   \includegraphics[width=1\linewidth]{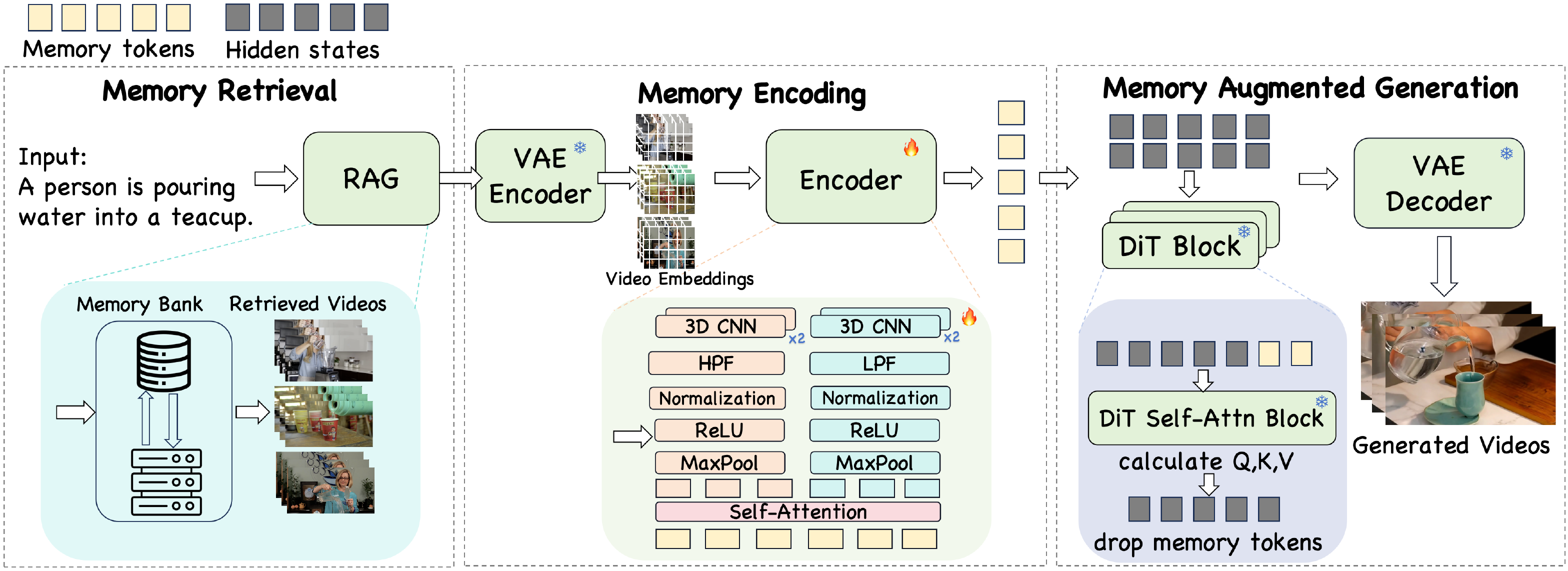}

\caption{\textbf{Overview of our DiT-Mem framework.}
Given a text prompt, we retrieve the top-$k$ relevant videos from a large external memory, encode them with the model's VAE to obtain video latents, and feed these latents into a memory encoder (3D CNNs, LPF/HPF filtering, and shared self-attention) to produce compact memory tokens.
During diffusion sampling, the memory tokens are concatenated with the video tokens before each self-attention layer of the frozen DiT backbone and participate in standard multi-head self-attention as queries, keys, and values.
We keep only the updated video tokens while reusing the same memory tokens at every layer, providing plug-and-play memory guidance for video generation.}

   \label{fig:model}
\end{figure*}

\section{Methodology}
\label{sec:method}
While Section~\ref{sec:empirical} analyzes how training-free steering vectors can intervene on physics-related features, directly optimizing cross-attention activations tends to interfere with text–video alignment.
Motivated by these observations, we propose a trainable, memory-augmented framework for DiT-based video generation models that improves physical fidelity while providing effective guidance to a frozen backbone.
Concretely, we introduce a lightweight memory module that can be paired with an off-the-shelf DiT.
We first present the overall DiT-Mem architecture and the design of the memory encoder, and then describe how to train the memory encoder end-to-end and how to perform plug-and-play, memory-guided inference.
The overview of our approach is shown in Figure~\ref{fig:model}.

\subsection{Memory Encoder}
We introduce a lightweight memory encoder that transforms each retrieved reference video into a compact, fixed-length sequence of memory token embeddings. These embeddings are concatenated with the hidden states of the frozen DiT to provide guidance.
The encoder is composed of 3D convolution layers for feature downsampling, low-pass and high-pass filters for frequency disentanglement, and self-attention layers for semantic aggregation.
\vspace{-1mm}
\paragraph{3D CNN for Downsampling.}
Given each retrieved video, we first convert it into visual feature maps using the VAE of the video generation model.
To condense the visual features into a small number of embeddings, we incorporate 3D convolutions with pooling layers to perform downsampling and extract spatio-temporal features.
The encoder consists of two stacked 3D convolutional blocks that reduce the spatial–temporal resolution of the visual feature maps while increasing the channel dimensionality.
Each 3D CNN block consists of a 3D convolution, followed by the frequency-domain filter module (detailed below), batch normalization, ReLU activation, and 3D max pooling.
After the convolutional computation, we flatten the spatial dimensions of the output and apply adaptive average pooling along the temporal axis to obtain a fixed-length sequence of token embeddings.
Each token is then projected into a shared memory embedding space via a linear transformation. 

\vspace{-1mm}
\paragraph{Filters for Disentanglement.}
In DiT-Mem, we apply low-pass and high-pass filters to disentangle the low-level appearance information and high-level dynamic cues. We first convert the visual features into the frequency domain, then multiple corresponding filters, and finally convert them back to the time domain.
Concretely, given a spatio-temporal feature tensor $x \in \mathbb{R}^{C \times D \times H \times W}$, we apply a 3D FFT along temporal and spatial dimensions:
\begin{equation}
\hat{x} = \mathcal{F}_{(D,H,W)}(x) \in \mathbb{C}^{C \times D \times H \times W}.
\end{equation}
We then apply two masks in the frequency domain, \ie a low-pass filter $\mathbf{M}_{\text{low}}$ and a high-pass filter $\mathbf{M}_{\text{high}}$, to isolate different frequency bands.
For $\mathbf{M}_{\text{low}}$ and $\mathbf{M}_{\text{high}}$, we attenuate the high-frequency and low-frequency components by scaling their magnitudes to $0.2$, respectively.
Thus, the low-pass and high-pass filtered features are obtained by element-wise multiplication in the frequency domain:
\begin{equation}
\hat{x}_{\text{low}} = \hat{x} \odot \textbf{M}_{\text{low}}, \quad
\hat{x}_{\text{high}} = \hat{x} \odot \textbf{M}_{\text{high}}.
\end{equation}
Next, the filtered signals are transformed back via inverse FFT. Following existing work~\cite{cai2021frequency,koo2024flexiedit}, we also add a residual connection here to stabilize training:
\begin{equation}
x_{\text{low}} = \mathcal{F}^{-1}(\hat{x}_{\text{low}})+x, \quad  x_{\text{high}} =\mathcal{F}^{-1}(\hat{x}_{\text{high}}) + x.
\end{equation}
This module is placed between each convolution and normalization layers. It helps to separate low-frequency appearance cues from high-frequency motion patterns, improving the disentanglement and supporting flexible use of the resulting memory tokens to guide inference.

\paragraph{Self-attention for Feature Aggregation.}
To capture long-range dependencies and enhance contextual representation, we apply lightweight self-attention modules to the resulting token embeddings. Specifically, we add a transformer-style self-attention block with shared parameters for the resulting low-frequency and high-frequency memory tokens:
\begin{equation}
x_{\text{attn}}^{\text{low}} = \text{SA}_{\phi}\!\left(x_{\text{mem}}^{\text{low}}\right), 
\qquad
x_{\text{attn}}^{\text{high}} = \text{SA}_{\phi}\!\left(x_{\text{mem}}^{\text{high}}\right),
\end{equation}
where $\text{SA}_{\phi}$ denotes the shared multi-head self-attention module.
The outputs are then concatenated along the token dimension to form the final memory sequence for the video.
This parameter sharing reduces overhead while enforcing consistent aggregation across appearance (low-pass) and motion (high-pass) streams.

\subsection{Memory-augmented DiT}
Based on the memory encoder, we can convert all retrieved videos into fixed-length visual token embeddings.
We directly insert them into the self-attention layers of the DiT and perform end-to-end training of the memory encoder to support plug-and-play memory-guided inference.
\vspace{-3mm}
\paragraph{Integrating into Self-attention Layers.}
To enable memory-guided video generation, we append the token embeddings produced by the memory encoder to the hidden states before each self attention layer of the DiT.
The resulting extended sequence, which contains both video tokens and memory tokens, is then fed into the standard multi-head self-attention, so that memory tokens participate as queries, keys, and values together with the video tokens.
After the attention operation, we discard the outputs at the memory positions and keep only the updated video tokens for the subsequent MLP and deeper layers, while the same set of memory tokens is re-injected before the next self-attention block.
This design conditions the frozen DiT backbone on external video-derived knowledge, while allowing gradients to flow to the memory encoder through the attention mechanism.
Since we insert only a small number of memory tokens (e.g., five reference videos leading to roughly 1{,}000 memory tokens) into the original input sequence (about 30{,}000 tokens), the additional attention cost is modest.
\vspace{-1mm}
\paragraph{End-to-end Memory Encoder Training.}
We train the memory encoder while keeping the DiT backbone frozen.
For each training sample, we retrieve $K$ reference videos, pass them through the plug-and-play memory encoder, and obtain memory tokens that guide the diffusion process.
Let $X_{\text{mem}}$ denote the sequence of memory tokens.
The denoising objective is the standard noise-prediction loss with memory conditioning:
\begin{equation}
\mathcal{L}_{\text{denoise}}
= \mathbb{E}_{x_t,\,\epsilon}\!\left[\;\big\|\epsilon-\epsilon_{\theta}\!\big(x_t,t,c,\,X_{\text{mem}}\big)\big\|_2^2\;\right].    
\end{equation}
Throughout training, the DiT backbone remains frozen; only the memory encoder is updated.
Since our memory encoder is lightweight, it is not data-hungry and about 10k samples are sufficient for training.
Thus, our training process is both parameter- and data-efficient, requiring only modest supervision to learn a generalizable plug-and-play memory aligned with the frozen DiT representation space
\vspace{-2mm}
\paragraph{Plug-and-play Memory-guided Inference.}
During inference, we follow the same retrieval$\rightarrow$encoding$\rightarrow$injection pipeline to guide video synthesis.
For each prompt, we retrieve the top-$K$ reference videos and run the trained memory encoder to obtain memory tokens, which are concatenated with the hidden states in the self-attention layers at every denoising step.
In practice, we pre-compute the token embeddings of all videos in the memory bank and store them in the database.
These embeddings can then be directly retrieved and injected in a plug-and-play manner, providing a flexible way to improve realism and physical consistency with negligible additional latency.

\subsection{Discussion}
In Table~\ref{tab:method_comparison}, we compare our method with eight closely related approaches for the video generation task. They can be roughly divided into three categories: retrieval-augmented generation (RAGME \citep{peruzzo2025ragmeretrievalaugmentedvideo}, MotionRAG \citep{zhumotionrag}, VideoRAG \citep{luo2024video}), diffusion memory methods (Corgi \citep{wu2025corgi}, VMem \citep{li2025vmem}, MALT \citep{yu2025malt}, WorldMem \citep{xiao2025worldmem}), and context compression techniques (LoViC \citep{jiang2025lovic}). The comparison covers two dimensions: Properties and Efficiency.
For properties, we consider whether the memory is plug-and-play, learnable, and contains high-level and low-level features.
For efficiency, we consider the training cost in terms of the data and trainable parameters.

While prior work has augmented video generation models with external context or memory, they typically rely on extensive training data and large-scale parameter updates. Many approaches also require fine-tuning the diffusion model, which hinders the portability and scalability. Others, while plug-and-play, often lack learnable components, limiting the adaptability across domains or tasks.

In contrast, our method introduces a lightweight and modular memory mechanism that is both plug-and-play and learnable, requiring only 10K training samples and a small number of tunable parameters (\eg 150M). Crucially, our memory operates independently of the base diffusion model, enabling seamless integration without retraining. This design allows our approach to generalize across scenarios with minimal cost. By supporting efficient memory conditioning without modifying the backbone, our work offers a unified and scalable framework for memory-augmented generation. 

\begin{table}[t]
\centering
\scriptsize                    
\caption{
\textbf{Comparison between our approach and prior works.}
\textbf{P\&P} denotes plug-and-play usage.
\textbf{Learn.} indicates whether the memory is learnable.
\textbf{HF./LF.} indicate support for high-/low-frequency guidance.
\textbf{Data} and \textbf{Para.} denote the amount of training data and the number of trainable parameters of the memory module.
}
\label{tab:method_comparison}
\resizebox{\columnwidth}{!}{
\begin{tabular}{l c c c c r r}
\toprule
& \multicolumn{4}{c}{\textbf{Properties}} & \multicolumn{2}{c}{\textbf{Efficiency}} \\
\cmidrule(lr){2-5} \cmidrule(lr){6-7}
\textbf{Method} & \textbf{P\&P} & \textbf{Learn.} & \textbf{HF.} & \textbf{LF.} & \textbf{Data} & \textbf{Para.} \\
\midrule

\multicolumn{7}{l}{\emph{Multimodal RAG}} \\
RAGME \citep{peruzzo2025ragmeretrievalaugmentedvideo} & $\times$ & $\times$ & $\checkmark$ & $\times$ & 10M & NA \\
MotionRAG \citep{zhumotionrag}                         & \checkmark & \checkmark & $\checkmark$ & $\times$ & 1M & 800M \\
VideoRAG \citep{luo2024video}                          & \checkmark & $\times$   & $\checkmark$ & $\checkmark$ & NA & NA \\
\midrule

\multicolumn{7}{l}{\emph{Diffusion Memory}} \\
Corgi \citep{wu2025corgi}       & $\times$ & $\times$   & $\times$ & $\checkmark$ & 100 & NA \\
VMem \citep{li2025vmem}         & \checkmark & $\times$   & $\times$ & $\checkmark$ & 10k & NA \\
MALT \citep{yu2025malt}         & $\times$ & \checkmark & $\checkmark$ & $\checkmark$ & 1B & 440M \\
WorldMem \citep{xiao2025worldmem} & $\times$ & \checkmark & $\checkmark$ & $\checkmark$ & 85K & NA \\
\midrule

\multicolumn{7}{l}{\emph{Context Compression}} \\
LoViC \citep{jiang2021focal} & $\times$ & \checkmark  & \checkmark & \checkmark & 1.6M & 2.3B \\
\midrule

\multicolumn{7}{l}{\emph{Ours}} \\
\textbf{DiT-Mem} & \checkmark & \checkmark & \checkmark & \checkmark & 10K & 150M \\
\bottomrule
\end{tabular}%
}
\end{table}

\section{Experiment}
\label{sec:experiment}

\subsection{Experimental Setup}

\paragraph{Baseline Methods.}

\begin{table*}[t]
\centering
\caption{\textbf{Comparison with baselines on PhyGenBench.}
Our models (Wan2.1+Ours and Wan2.2+Ours) achieve the best performance within their respective backbone groups across almost all physics domains. BEST results in each group are highlighted in bold.}
\label{tab:phygen}
\small
\begin{tabular}{l l c c c c c c}
\toprule
\textbf{Source} & \textbf{Model} & \textbf{Size} &
\textbf{Mechanics}($\uparrow$) & \textbf{Optics}($\uparrow$) &
\textbf{Thermal}($\uparrow$) & \textbf{Material}($\uparrow$) &
\textbf{Average}($\uparrow$) \\
\midrule
\multirow{3}{*}{Closed-source}
& Pika \citep{pika2024}                & --   & $0.35$ & $0.56$ & $0.43$ & $0.39$          & $0.44$ \\
& Gen-3 \citep{runway2024gen3}         & --   & $\mathbf{0.45}$ & $0.57$ & $0.49$ & $\mathbf{0.51}$ & $\mathbf{0.51}$ \\
& Kling \citep{kling2024}              & --   & $\mathbf{0.45}$ & $\mathbf{0.58}$ & $\mathbf{0.50}$ & $0.40$          & $0.49$ \\
\midrule
\multirow{9}{*}{Open-source}
& CogVideoX \citep{yang2024cogvideox}  & 2B   & $0.38$ & $0.43$ & $0.34$ & $0.39$ & $0.39$ \\
& CogVideoX                             & 5B   & $0.39$ & $0.55$ & $0.40$ & $0.42$ & $0.45$ \\
& Open-Sora V1.2 \citep{opensora2024}  & 1.1B & $0.43$ & $0.50$ & $0.44$ & $0.37$ & $0.44$ \\
& Lavie \citep{gupta2023lavie}         & 860M & $0.30$ & $0.44$ & $0.38$ & $0.32$ & $0.36$ \\
& Vchitect 2.0 \citep{vchitect2024}    & 2B   & $0.41$ & $0.56$ & $0.44$ & $0.37$ & $0.45$ \\
\cmidrule{2-8}
& Wan2.1 \citep{wan2024video} & 1.3B & $0.48$ & $0.67$ & $0.38$ & $0.37$ & $0.48$ \\
& \textbf{Wan2.1+Ours}            & 1.3B & $0.58$ & $0.73$ & $0.47$ & $0.39$ & $0.54$ \\
\cmidrule{2-8}
& Wan2.2 \citep{wan2025wan}   & 5B   & $\mathbf{0.60}$ & $0.69$ & $0.42$ & $0.45$ & $0.54$ \\
& \textbf{Wan2.2+Ours}            & 5B   & $0.56$ & $\mathbf{0.74}$ & $\mathbf{0.48}$ & $\mathbf{0.47}$ & $\mathbf{0.56}$ \\
\bottomrule
\end{tabular}
\end{table*}

\begin{table*}[t]
\centering
\caption{\textbf{Comparison on VBench.} The evaluation separates metrics into low-level (appearance/texture) and high-level (semantic/dynamic) categories. Our method maintains competitive performance across both groups. The Wan2.2+Ours variant achieves the strongest low-level and high-level average scores among all compared models. BEST results in each group are highlighted in bold.}
\label{tab:vbench}
\small
\setlength{\tabcolsep}{4pt}
\begin{tabular}{l c c c c c c c c c}
\toprule
\multirow{2}{*}{\textbf{Models}} &
\multicolumn{4}{c}{\textbf{Low-level Metrics}} &
\multicolumn{5}{c}{\textbf{High-level Metrics}} \\
\cmidrule(lr){2-5} \cmidrule(lr){6-10}
& \textbf{Color}
& \textbf{Style}
& \textbf{Obj. Class}
& \textbf{Low-Avg}
& \textbf{Subj. Consist.}
& \textbf{Action}
& \textbf{Spatial Rel.}
& \textbf{Multi-Obj.}
& \textbf{High-Avg} \\
\midrule
OpenSora V1.1~\cite{opensora2024}
    & $74.56\%$ & $\mathbf{23.50\%}$ & $86.76\%$ & $61.61\%$
    & $92.35\%$ & $84.20\%$ & $52.47\%$ & $40.97\%$ & $67.00\%$ \\

ModelScope~\cite{modelscope2023t2v}
    & $81.72\%$ & $23.39\%$ & $82.25\%$ & $62.45\%$
    & $89.87\%$ & $92.40\%$ & $33.68\%$ & $38.98\%$ & $63.23\%$ \\

VideoCrafter~\cite{chen2023videocrafter}
    & $78.84\%$ & $21.57\%$ & $\mathbf{87.34\%}$ & $62.58\%$
    & $86.24\%$ & $\mathbf{93.00\%}$ & $36.74\%$ & $25.93\%$ & $60.48\%$ \\

CogVideo~\cite{hong2022cogvideo}
    & $79.57\%$ & $22.01\%$ & $73.40\%$ & $58.33\%$
    & $92.19\%$ & $78.20\%$ & $18.24\%$ & $18.11\%$ & $51.18\%$ \\
\midrule

Wan 2.1
    & $91.02\%$ & $19.91\%$ & $74.87\%$ & $61.93\%$
    & $\mathbf{96.96\%}$ & $80.00\%$ & $64.19\%$ & $59.12\%$ & $75.07\%$ \\

\textbf{Wan 2.1+Ours}
    & $88.96\%$ & $20.05\%$ & $64.63\%$ & $57.88\%$
    & $96.29\%$ & $82.00\%$ & $71.35\%$ & $61.00\%$ & $77.16\%$ \\
\midrule

Wan 2.2
    & $85.92\%$ & $21.26\%$ & $79.12\%$ & $62.10\%$
    & $95.51\%$ & $82.00\%$ & $\mathbf{78.91\%}$ & $69.12\%$ & $81.39\%$ \\

\textbf{Wan 2.2+Ours}
    & $\mathbf{93.67\%}$ & $21.22\%$ & $80.00\%$ & $\mathbf{64.96\%}$
    & $95.67\%$ & $88.00\%$ & $78.42\%$ & $\mathbf{74.38\%}$ & $\mathbf{84.12\%}$ \\
\bottomrule
\end{tabular}
\end{table*}

We compare our approach against a broad set of state-of-the-art video generation systems, including both open-source and commercial models. The open-source baselines span diverse architectures such as DiT- and U-Net–based diffusion models and 3D VAEs, including CogVideo~\cite{hong2022cogvideo}, ModelScope~\cite{modelscope2023t2v}, VideoCrafter~\cite{chen2023videocrafter}, LaVie~\cite{gupta2023lavie}, Open-Sora~\cite{opensora2024}, Vchitect~\cite{vchitect2024}, CogVideoX~\cite{yang2024cogvideox}, Wan2.1~\cite{wan2024video} and Wan2.2~\cite{wan2025wan}. We additionally benchmark against competitive proprietary systems—Pika~\cite{pika2024}, Runway Gen-3~\cite{runway2024gen3}, and Kling~\cite{kling2024}. Together, these baselines represent the strongest publicly available and commercial T2V models and provide a comprehensive evaluation of our method. A detailed description of each baseline is provided in the Supplementary Material.

\paragraph{Evaluation Benchmarks.}

We assess our method on three suites that cover physics compliance, per-dimension visual quality, and standard T2V metrics: (1) \textbf{PhyGenBench}~\cite{meng2024worldsimulatorcraftingphysical} for physical commonsense, evaluated under the official protocol with GPT-4o as the MLLM judge; (2) \textbf{VBench}~\cite{huang2024vbench}, following the per-dimension protocol on a pre-registered subset balancing temporal and perceptual/semantic factors; and (3) \textbf{OpenVidHD Test Set~\citep{nan2025openvid1mlargescalehighqualitydataset}} (100 randomly sampled clips, seed fixed) with standard metrics including FVD, CLIP, and Action/Motion scores. All methods share identical prompts, seeds, and inference settings. Full protocols, prompts, scoring pipelines, and additional breakdowns are in appendix.

\subsection{Main Results}
\label{subsec:result}
On PhyGenBench, augmenting the Wan backbones with our DiT-Mem module yields consistent improvements across all four physics domains, culminating in a new state-of-the-art average performance (Table~\ref{tab:phygen}). The DiT-Mem-1.3B (Wan2.1+Ours) not only achieves a substantial boost over its underlying baseline but also surpasses all existing open-source and several commercial systems, reaching performance levels comparable to the larger Wan2.2 TI2V-5B model. Building on this foundation, the DiT-Mem-5B (Wan2.2+Ours) further advances results and delivers state-of-the-art performance across most evaluation dimensions. Gains are particularly pronounced in optics and thermal reasoning, suggesting that the proposed memory encoder effectively retrieves and injects physics-relevant cues. This guidance enables the DiT backbone to better integrate dynamic and appearance information, resulting in more physically faithful video generation.

VBench emphasizes appearance quality and semantic correctness. Following its structure, we group the metrics into low-level (Color, Appearance Style, Object Class) and high-level (Subject Consistency, Human Action, Spatial Relationship, Multiple Objects) categories. 
Subject consistency remains on par with the backbone models, and appearance style is largely preserved, indicating that the memory module does not disrupt visual aesthetics. A mild trade-off in object-class accuracy is observed on the smaller backbone, but this effect diminishes on the DiT-Mem-5B (Wan2.2+Ours), consistent with our design goal of prioritizing dynamic and relational cues.
The memory mechanism provides complementary improvements on VBench without degrading image quality. Our DiT-Mem-1.3B variant remains competitive and DiT-Mem-5B (Wan2.2+Ours) configuration achieves the highest low-level and high-level average scores among all compared models. 
Overall, our plug-and-play DiT-Mem methods complement backbone capacity and delivers state-of-the-art VBench averages without compromising visual aesthetics.

\begin{table*}[t]
\centering
\caption{\textbf{Frequency ablation on VBench with metrics grouped into low- and high-level categories.} LPF-only performs better on appearance-related metrics, while HPF-only excels on semantic/dynamic ones, validating the complementary roles of the two frequency filters.}
\label{tab:freq_study}
\small
\setlength{\tabcolsep}{6pt}
\renewcommand{\arraystretch}{0.95}

\begin{tabular}{l c c c c c c c}
\toprule
\multirow{2}{*}{\textbf{}} &
\multicolumn{3}{c}{\textbf{Low-level Metrics}} &
\multicolumn{4}{c}{\textbf{High-level Metrics}} \\
\cmidrule(lr){2-4} \cmidrule(lr){5-8}
& \textbf{Color}
& \textbf{Style}
& \textbf{Obj. Class}
& \textbf{Subj. Consist.}
& \textbf{Action}
& \textbf{Spatial Rel.}
& \textbf{Multi-Obj.} \\
\midrule

\textbf{Wan2.1 + Ours}
    & 88.96\% & 20.05\% & 64.63\%
    & 96.29\% & 82.00\% & 71.35\% & 61.00\% \\
\midrule

\textbf{w/o HPF (LPF only)}
    & 87.72\% & 20.24\% & 67.63\%
    & 96.67\% & 82.00\% & 61.50\% & 65.50\% \\

\textbf{w/o LPF (HPF only)}
    & 80.03\% & 20.34\% & 67.25\%
    & 96.15\% & 82.00\% & 73.03\% & 75.13\% \\
\bottomrule
\end{tabular}
\end{table*}

Beyond the two benchmarks, we also include standard T2V metrics for completeness (Table~\ref{tab:low-level}). 
Our DiT-Mem method attains \emph{lower FVD} and \emph{slightly higher} Action/CLIP, while keeping motion magnitude competitive. 
In short, our method improves temporal coherence without degrading common alignment or recognizability metrics.

\begin{table}[t]
  \centering
  \small
  \caption{\textbf{Evaluation on the Open-Vid test set using standard low-level video generation metrics.} Wan2.1+Ours maintains strong text–video alignment, action accuracy, and motion quality, while achieving the lowest FVD, indicating that the physics-oriented improvements of our method do not degrade general video quality.}
  \label{tab:low-level}
  \setlength{\tabcolsep}{3pt}      %
  \renewcommand{\arraystretch}{0.95} %
  \begin{tabular*}{\columnwidth}{@{\extracolsep{\fill}} l c c c c @{}}
    \toprule
    \textbf{Model} & \textbf{Action}($\uparrow$) & \textbf{CLIP}($\uparrow$) & \textbf{FVD}($\downarrow$) & \textbf{Motion}($\uparrow$) \\
    \midrule
    CogVideoX-5B    & 45.09 & 48.38 & 1420.56 & 3.85 \\
    VideoCrafter2   & 42.35 & 47.01 & 1842.92 & \textbf{4.89} \\
    Wan2.1    & 48.35 & 60.48 & 1479.10 & 3.90 \\
    \textbf{Wan 2.1+Ours}   & \textbf{48.72} & \textbf{60.56} & \textbf{1354.22} & 4.18 \\
    \bottomrule
  \end{tabular*}
\end{table}

\subsection{Further Analysis}

\paragraph{Frequency Control}
The high-pass and low-pass filters play a central role in our memory encoder. By design, these filters aim to disentangle low-level appearance cues from high-level semantic and dynamic information, enabling the DiT backbone to more effectively leverage external world knowledge during both training and inference. To understand what types of information each frequency branch captures, we perform a controlled study on VBench using the same low-level and high-level metric grouping introduced in the Sec.~\ref{subsec:result}. The results are shown in Table~\ref{tab:freq_study}. The findings clearly demonstrate complementary behaviors between the two branches. The model using only the low-pass filter achieves higher scores on most low-level metrics, indicating its strength in capturing appearance, identity, and color-related cues that correspond to low-frequency signals. In contrast, the high-pass-only variant consistently outperforms on high-level metrics, suggesting that high-frequency components are more informative for capturing semantic structure, spatial relations, and subtle dynamic patterns. These results strongly validate our hypothesis.
\subsection{Ablation Study}
\label{ablation}
\begin{table*}
\centering
\small
\caption{
\textbf{Ablation study on DiT-Mem-1.3B evaluating the contribution of each component of the proposed memory encoder.} We report variants obtained by progressively adding the 3D convolution layers (3D), high-pass filter (HPF), low-pass filter (LPF), and attention modules—either separate attention (SPA) or shared attention (SA).
}
\begin{tabular}{lccccc}
\toprule
\textbf{Variant} &
\textbf{Force}($\uparrow$) &
\textbf{Optics}($\uparrow$) &
\textbf{Thermal}($\uparrow$) &
\textbf{Material}($\uparrow$) &
\textbf{Avg}($\uparrow$) \\
\midrule
Baseline (Wan2.1-T2V-1.3B)                & 0.48 & 0.67 & 0.38 & 0.37 & 0.47 \\
+ 3D                       & 0.53 & 0.69 & 0.42 & 0.37 & 0.50 \\
+ 3D + HPF                 & 0.48 & 0.67 & 0.40 & 0.41 & 0.49 \\
+ 3D + HPF + LPF           & 0.48 & 0.69 & 0.43 & 0.39 & 0.50 \\
+ 3D + HPF + LPF + SPA  & 0.52 & 0.71 & 0.38 & \textbf{0.42} & 0.51 \\
\textbf{+ 3D + HPF + LPF + SA (Ours)} 
                         & \textbf{0.58} & \textbf{0.74} & \textbf{0.47} & 0.39 & \textbf{0.54} \\
\bottomrule
\end{tabular}
\label{tab:architecture}
\end{table*}

\begin{figure*}
    \centering
    \includegraphics[width=0.9\linewidth]{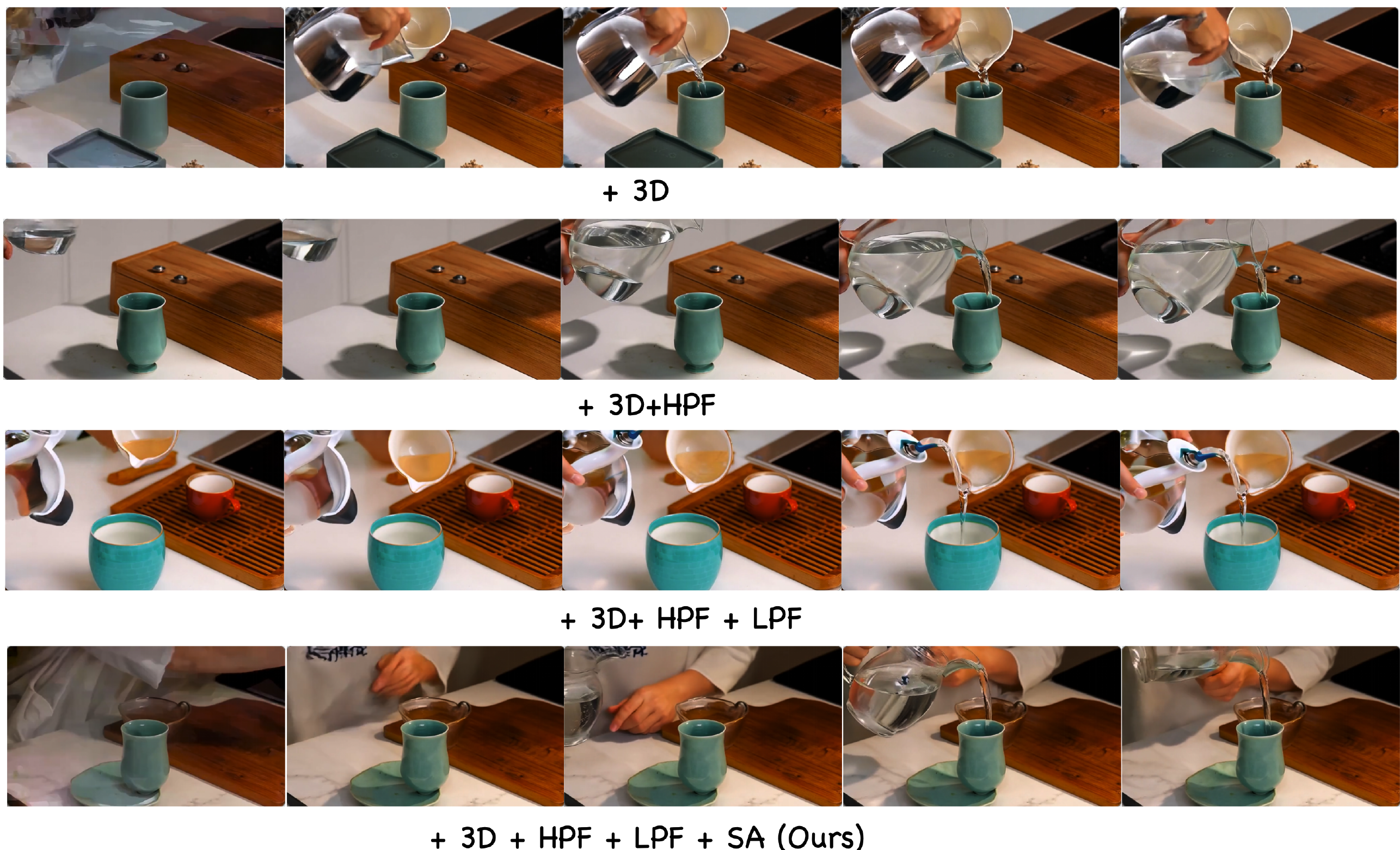}
    \caption{\textbf{Visual ablation study.} All variants are generated using the same text prompt (``a person is pouring water into a teacup''), same seed, and identical set of five retrieved reference videos. The baseline with 3D convolution layers (+3D) suffers from severe artifacts, appearing as a hallucinated second floating cup. Adding the high-pass filter (+HPF) resolves this structural issue and improves motion, but results in blurred details on the person and hand. While incorporating the low-pass filter (+LPF) introduces appearance features from other objects, our full model with Shared Attention (SA) achieves the best balance. It effectively enhances motion without over-injecting retrieved object appearances, thereby preserving the semantic fidelity of the original text prompt.}
    \label{fig:diff_architec}
\end{figure*}

To understand the contribution of each component in our memory encoder, we perform an ablation study on PhyGenBench by progressively adding the 3D convolution layers, frequency filters, and attention module (Table \ref{tab:architecture}). Qualitative comparisons are shown in Figure \ref{fig:diff_architec}.

Starting from the baseline, adding 3D convolution layers yields a clear performance gain in extracting spatiotemporal features. However, as shown in Figure \ref{fig:diff_architec} (Row 1), this variant can suffer from severe structural hallucinations, manifested as an unphysical second floating cup.

Introducing the high-pass filter (HPF) successfully resolves this structural artifact and improves motion dynamics (Row 2). Yet, relying predominantly on high-frequency cues introduces a trade-off: it tends to blur fine-grained details of the subject (e.g., the person and hand), as it filters out the semantic richness required for sharp subject representation.

Incorporating the low-pass filter (LPF) branch addresses this by introducing essential appearance information from the retrieved context, stabilizing the object's visual coherence (Row 3). Finally, the shared-attention (SA) module provides the most significant improvement. By allowing high- and low-frequency streams to interact, our full model achieves an optimal balance: it enhances physical motion without over-injecting the specific appearances of retrieved objects. This ensures that the generated video remains semantically faithful to the text prompt while exhibiting realistic fluid dynamics and geometry.

\section{Related Work}
\label{sec:related}

\paragraph{Diffusion Models for Video Generation} Video generation has advanced along two main lines: U-Net denoisers and Transformer sequence models. U-Net models, such as Stable Video Diffusion \citep{blattmann2023stable} and Lumiere \citep{bar2024lumiere}, build on LDMs \citep{rombach2022high} using space-time U-Nets for temporal consistency \cite{blattmann2023align,wang2023modelscope,he2022latent}. A newer paradigm, the Diffusion Transformer \citep{DiT}, replaces the U-Net with a Transformer over latent patches for better scaling, as adopted by leading models \citep{wan2025wan,yang2024cogvideox,kong2024hunyuanvideo}. Despite these advances, a persistent physics gap remains—scaling alone does not yield reliable physical reasoning—motivating explicit mechanisms to inject structured world knowledge \citep{meng2024towards,motamed2025generative,huang2024vbench}.

\paragraph{Memory for Generative Models}
Memory mechanisms inject external information into frozen generators to overcome limited world knowledge and long-range context without full retraining. For images, RDM, kNN-Diffusion, and Re-Imagen condition on nearest neighbors to improve fidelity but depend on large corpora and incur additional latency~\cite{blattmann2022retrieval,sheynin2022knn,chen2022re}. RAGME~\citep{peruzzo2025ragmeretrievalaugmentedvideo} retrieves reference videos to enhance motion realism but still requires minimal fine-tuning of the diffusion modle, while MotionRAG~\citep{zhumotionrag} uses a context-aware motion adapter and motion-injection adapter, relying on heavy adapter stacks and lacking explicit disentanglement. Inspired by the low-latency, parametric plug-and-play goal of VLM memory modules~\citep{wenyitowards}, we instead introduce a compact, trainable video memory encoder with explicit frequency-based disentanglement of appearance and dynamics, enabling truly plug-and-play, low-latency guidance without backbone updates.

\paragraph{Frequency-Aware Representations}
Frequency-aware representations separate global structure from fine detail. Focal Frequency Loss emphasizes hard-to-synthesize spectral components~\cite{jiang2021focal}, and filter-enhanced MLPs show that learnable frequency masks can denoise sequences and capture salient patterns~\cite{zhou2022filter}; in generative modeling, Free-T2M decouples low and high frequency motion to improve semantic alignment \cite{chen2025free}. Recent work also applies frequency decomposition in text-to-video generation: ConsisID~\cite{yuan2025identity} preserves human identity by separating low-frequency global structure from high-frequency facial details. However, most frequency methods target image-to-image translation or style transfer—such as FDIT~\citep{cai2021frequency} and FreMixer~\citep{li2023frequency}—rather than controllable, frequency-guided conditioning. We instead embed learnable high/low-pass filters into a plug-in memory encoder, disentangling appearance and dynamics in retrieved references to guide DiT-based video generation.

\section{Conclusion}
\label{sec:conclusion}
In this work, we equipped DiT-based video diffusion models with a world knowledge memory to reduce violations of physics and commonsense. Starting from empirical evidence that DiT can be steered through interventions on its hidden states, we showed that simple low-pass and high-pass filtering in the embedding space can disentangle low-level appearance cues from high-level physical and semantic information, enabling targeted guidance during generation. Building on these findings, we introduced a learnable, plug-and-play memory encoder DiT-Mem that converts reference videos into compact memory tokens via stacked 3D CNNs, frequency-aware filters, and self-attention, and injects them as in-context memory into the DiT self-attention layers.
Crucially, our design keeps the diffusion backbone frozen and only trains the memory encoder, resulting in a parameter-efficient and data-efficient solution that is easy to deploy across different DiT-based video generators. Experiments on various benchmarks and models demonstrate the effectiveness of our memory-guided approach.

{
    \small
    \bibliographystyle{ieeenat_fullname}
    \bibliography{main}
}
\clearpage
\setcounter{page}{1}
\maketitlesupplementary
\appendix

\section{Overview}
\label{sec:overview}
In this supplementary, we first provide the implementation details of the proposed method (Sec.~\ref{sec:implementation}). Next, we provide extended descriptions of all baseline models used in our comparisons (Sec.~\ref{sec:baselines_supp}) and detailed specifications of the evaluation benchmarks (Sec.~\ref{benchmarks}). We then analyze the impact of memory size on PhyGenBench performance using DiT-Mem-1.3B, which is built on the Wan2.1 backbone (Sec.~\ref{Memory_Size}). Finally, we provide a comprehensive case study comparing our DiT-Mem method against baseline models across diverse scenarios. (Sec. ~\ref{Case_Study})

\section{Implementation Details}
\label{sec:implementation}
Our pipeline has three stages: \emph{video retrieval}, \emph{memory encoding}, and \emph{memory-augmented generation}. Unless otherwise noted, all experiments use the top-5 retrieved videos for fairness across settings. 
We construct our external memory bank from the OpenVidHD-0.4M captioned video dataset. All captions are encoded into dense vectors using the GTE-v1.5 model \citep{zhang2024mgte} and indexed with FAISS \citep{johnson2019billion}. At inference, the input prompt is embedded and the top matches are retrieved by inner-product similarity. 

During fine-tuning, we freeze the entire DiT backbone and optimize only the proposed memory encoder. We adopt a 10K subset of the OpenVidHD-0.4M dataset as the training split, while the remaining 430K videos are used exclusively to construct the retrieval index. For the low-level video quality evaluation in Benchmark~(3), we additionally sample 100 clips from OpenVidHD-0.4M, ensuring that they are strictly disjoint from both the training subset and the retrieval memory.

We fine-tune the memory encoder on two backbone models: Wan2.1-T2V-1.3B and Wan2.2-TI2V-5B. The learning rates are set to $1\times 10^{-5}$ for Wan2.1 and $5\times 10^{-5}$ for Wan2.2. For both settings, we use a batch size of 4 with 16 gradient-accumulation steps. All experiments are conducted on 8 NVIDIA A100 GPUs. Training runs for 1{,}500 steps on Wan2.1 and 1200 steps on Wan2.2. The resulting models are denoted as DiT-Mem-1.3B (built on Wan2.1) and DiT-Mem-5B (built on Wan2.2).

For evaluation, we use fixed seed at 42 in PhyGenBench and OpenVidHD Test Set. For VBench, each prompt requires five generated videos to assess consistency and stability. 
To ensure reproducibility, we assign deterministic seeds to the five videos of each prompt using
\[
\text{seed} = 42 + \text{prompt}_{\text{idx}} \times 10 + \text{video}_{\text{idx}},
\]
where \(\text{prompt}_{\text{idx}}\) is the prompt index and \(\text{video}_{\text{idx}} \in \{0,\ldots,4\}\) denotes the video within the prompt.

\section{Baseline Methods}
\label{sec:baselines_supp}
We compare our method against a series of open-source and proprietary video generation models. Among open-source baselines, (1) CogVideo~\cite{hong2022cogvideo} proposes a DiT-based video diffusion model with improved text–video alignment through hierarchical training.
(2) ModelScope~\cite{modelscope2023t2v} subsequently adopts a 3D U-Net diffusion architecture for large-scale T2V synthesis. (3) VideoCrafter~\cite{chen2023videocrafter} focuses on controllable image-to-video (I2V) generation using latent diffusion models with temporal conditioning. (4) Lavie~\cite{gupta2023lavie} further enhances concept compositionality through joint image-video fine-tuning. (5) Open-Sora V1.2~\cite{opensora2024} integrates a 3D VAE, rectified flow, and score conditioning to improve temporal consistency and visual quality. (6) Vchitect 2.0~\cite{vchitect2024} introduces a hybrid parallelism framework to jointly capture spatial and temporal dependencies. More recently, (7) CogVideoX~\cite{yang2024cogvideox} develops a 3D VAE and expert Transformer architecture to enable coherent long-duration video generation. (8) Wan2.1~\cite{wan2024video} and (9) Wan2.2~\cite{wan2025wan} represent large-scale video foundation models trained on extensive proprietary datasets, featuring advanced VAE architectures for high fidelity and stability. 

For proprietary models, we include Pika~\cite{pika2024}, Gen-3~\cite{runway2024gen3} and Kling~\cite{kling2024}, which are among the most competitive commercial video generation systems. (1) Pika is developed by Pika Labs, offering high-quality, text-driven video synthesis. (2) Gen-3 Alpha, released by Runway, is the third-generation video foundation model designed for creative and cinematic generation. (3) Kling AI, developed by Kuaishou, focuses on high-fidelity, real-world motion generation optimized for consumer applications.

\section{Evaluation Benchmarks}
\label{benchmarks}
We evaluate our methods on three comprehensive benchmarks that focus on different dimensions:

\textbf{(1) PhyGenBench.}
PhyGenBench targets physical commonsense in text-to-video generation, providing 160 prompts that span 27 physical laws across the domains of mechanics, optics, thermal phenomena, and material behavior, enabling quantitative assessment of adherence to real-world dynamics. It comprises three evaluation stages using multimodal large language models (MLLMs): a single-frame VQA test, a multi-frame physical-order verification, and a video-level naturalness score. We adopt the official protocol, using OpenAI GPT-4o as the MLLM.

\textbf{(2) VBench.}
VBench is a comprehensive, hierarchical benchmark for video generation that decomposes “quality” into well-defined dimensions for fine-grained, objective evaluation. Each dimension includes curated prompt suites and automated scoring pipelines, complemented by human-preference annotations that align well with the metrics.
We follow VBench’s per-dimension protocol and evaluate a pre-registered subset balancing temporal and perceptual/semantic factors. Table~\ref{tab:vbench} reports the per-dimension results. All methods use identical prompts, seeds, and inference settings, and scoring follows the official VBench implementations.

\textbf{(3) OpenVidHD Test Set.}
Beyond the first two benchmarks, we further evaluate our method using a set of widely adopted low-level video generation metrics. Specifically, we sample a random subset of 100 clips from the OpenVidHD-0.4M dataset—ensuring that none of these clips appear in either our memory bank or training data—and compute several standard metrics. These include \textbf{Fréchet Video Distance (FVD) \citep{unterthiner2019fvd}} for overall video quality, \textbf{CLIP} \citep{radford2021learning} for text video semantic alignment, and \textbf{Action/Motion} scores that quantify action recognizability and the magnitude and smoothness of motion. This evaluation suite verifies that the physics-oriented improvements introduced by our method do not come at the expense of general video generation quality, as summarized in Table~\ref{tab:low-level}.

\section{Memory Size Study}
\label{Memory_Size}
\begin{figure}[t]
  \centering
   \includegraphics[width=0.8\linewidth]{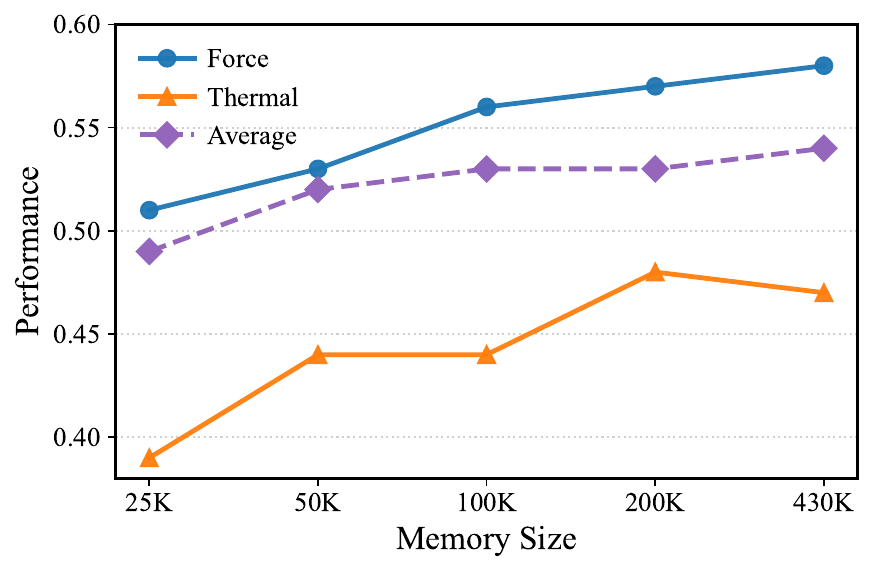}

   \caption{\textbf{Effect of memory size on PhyGenBench performance using DiT-Mem-1.3B}. Larger memory banks provide richer reference knowledge and yield consistently higher scores, while the model retains strong robustness even with significantly reduced memory capacity.}
   \label{fig:memory_size}
\end{figure}

The main component of our framework is the memory module, which stores video-derived world knowledge. Given an input caption, its embedding is used to retrieve the most relevant memory entries that guide the generation process. To understand how memory capacity affects performance, we evaluate models with different memory sizes on PhyGenBench and report the results in Fig.~\ref{fig:memory_size}. We observe a clear trend: larger memory sizes consistently lead to better performance across almost all metrics, as the full 430k memory provides a richer set of reference videos, enabling the memory encoder to extract more informative features and offer stronger guidance to the diffusion model. As the memory size decreases, the retrieval space shrinks and performance degrades. Nevertheless, our method remains surprisingly robust: even when the memory size is reduced to 1/20 of the full set, the model retains strong performance, which we attribute to the discriminative capability of the memory encoder in extracting high-quality multi-dimensional features while suppressing noise, demonstrating that our memory mechanism is both effective and efficient.

\section{Case Study}
\label{Case_Study}
\begin{figure*}
    \centering
    \includegraphics[width=0.8\linewidth]{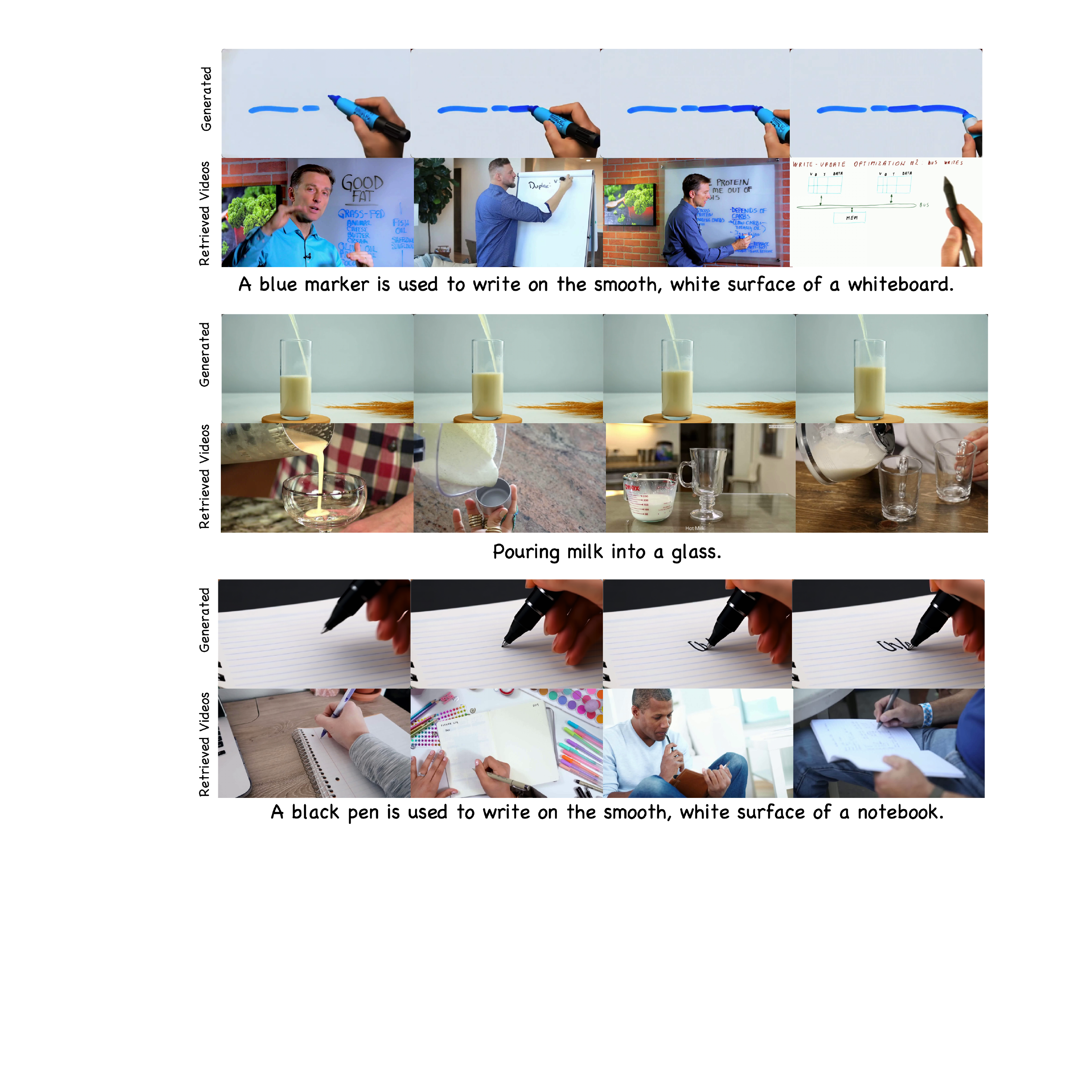}
    \caption{\textbf{Retrieval-augmented generation examples.} For each prompt, the top row displays sample frames from the video generated by our DiT-Mem-1.3B, while the bottom row shows the first frames of the corresponding retrieved reference videos. The results demonstrate that our model effectively retrieves and utilizes relevant information by leveraging the semantic alignment between the input prompt and video captions.}
    \label{fig:RAG_Samples}
\end{figure*}

As shown in Figure \ref{fig:RAG_Samples}, by leveraging the semantic alignment between the input prompt and reference video captions, our model effectively retrieves and utilizes relevant dynamic cues from diverse contexts. 

\begin{figure*}[ht] 
    \centering

    \begin{minipage}[t]{0.75\textwidth} 
        \centering
        \includegraphics[width=\linewidth]{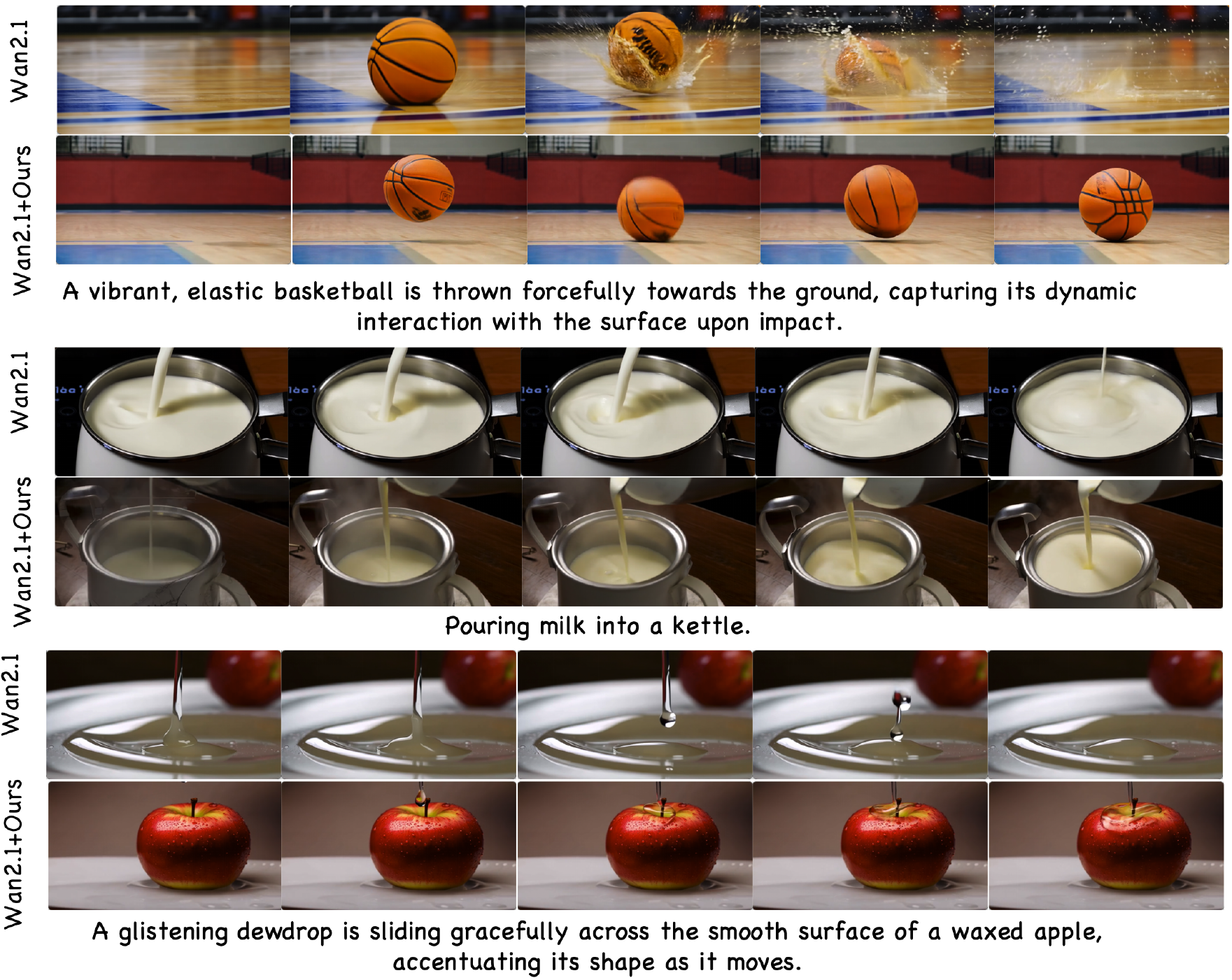}
        \label{fig:left_image_star}
    \end{minipage}
    \vspace{-5mm}

    \begin{minipage}[t]{0.75\textwidth} 
        \centering
        \includegraphics[width=\linewidth]{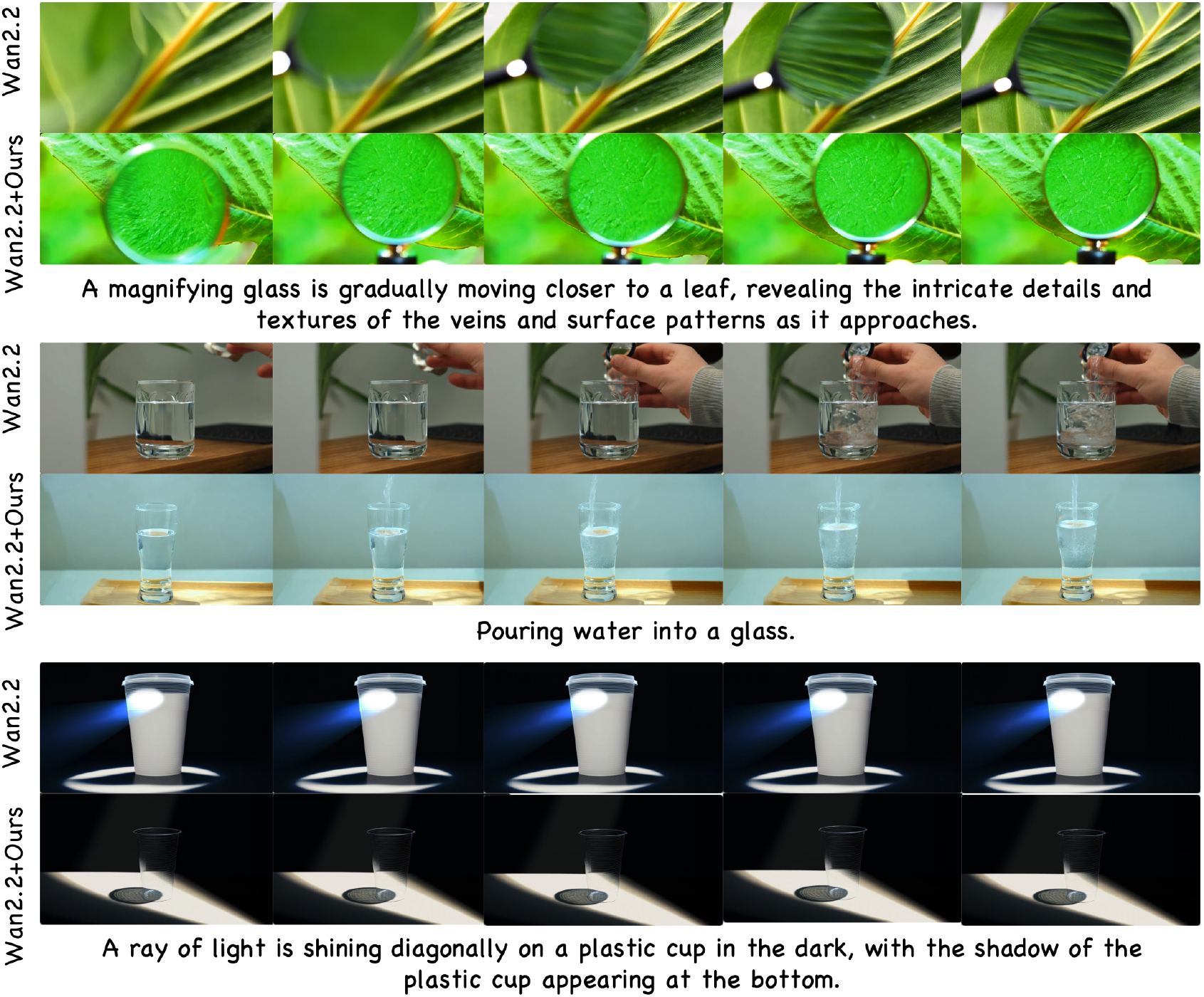}
        \label{fig:right_image_star}
    \end{minipage}
    \vspace{-5mm}
    \caption{\textbf{Qualitative comparison between the baseline models (Wan2.1/Wan2.2) and our method.} While the baselines occasionally exhibit physical hallucinations—such as the "liquid-like" splashing of a solid basketball (top-left) or missing shadows (bottom-right)—our method correctly models rigid body dynamics, fluid interactions, and lighting geometry, resulting in superior realism.} 
    \label{fig:Case_Study}
\end{figure*}

Figure \ref{fig:Case_Study} compares our proposed approach against baseline models (Wan2.1 and Wan2.2) across diverse physical scenarios. The results show consistent improvements in physical plausibility.

When enhancing Wan2.1, our method corrects fundamental violations of physical laws. For example, it transforms the unphysical "liquid-like splashing" of the basketball upon impact into a realistic rigid-body bounce. It also refines the dynamics of fluids, instilling natural flow patterns in the pouring milk and sliding dewdrop sequences, which appear distorted or defy gravity in the baseline.

Also, our approach induces accurate environmental interactions in Wan2.2 for the static objects, such as the plastic cup, which lacks proper cast shadows in the baseline video. It also corrects optical inconsistencies, replacing the unnatural rendering of the magnifying glass with physically grounded visual distortions.

\end{document}